\documentclass[letterpaper, 10 pt, conference]{ieeeconf}
\IEEEoverridecommandlockouts
\usepackage{cite}
\usepackage{amsmath,amssymb}
\usepackage{graphicx}
\usepackage{booktabs}
\usepackage{xcolor}
\usepackage[T1]{fontenc}
\usepackage[utf8]{inputenc}
\usepackage{textcomp}
\usepackage{url}
\usepackage{microtype}

\begin{document}

\title{\LARGE \bf Recovering Aggressively Pruned Vision-Language-Action\\
Models with Offline Hidden-State Distillation}

\author{Chiyoung Kim, Sanghyuk Roy Choi, and Minhyeok Lee$^{*}$%
\thanks{The authors are with Chung-Ang University, Seoul, Republic of Korea.
{\tt\small \{kimcy0829, choiroy, mlee\}@cau.ac.kr}}%
\thanks{$^{*}$Corresponding author.}%
}

\maketitle
\thispagestyle{empty}
\pagestyle{empty}

\begin{abstract}
Vision-language-action (VLA) models let robots follow language instructions,
but their language backbones of several billion parameters are the main
obstacle to running them on robot hardware. Structured pruning reduces that backbone, and removing 63\% of it from
OpenVLA-OFT drops LIBERO-Long success from 93.2\% to 0.8\%. A recent approach
restores such a model with supervised fine-tuning followed by reinforcement
learning, which needs online rollouts and hundreds of GPU-hours. We recover most of the lost
success entirely offline. Width pruning narrows the blocks but keeps the residual stream at its original
size, so teacher and student hidden states have the same shape and are matched
directly, without a projector. Training against a cache built in one teacher
pass lifts the 63\%-reduced student to within 3.5 points of the teacher in
about 8 GPU-hours. A sweep over nine ratios locates where the recovery objective starts to
matter. Up to 45\% reduction the two do not differ significantly on OpenVLA-OFT.
Hidden-state distillation then adds $+2.1$ to $+4.5$ points there between
63\% and 87\%, and $+9.4$ to $+22.1$ points on CogACT from 63\% onward. At 81\% on CogACT, a tripled recovery budget narrows the
distilled student's gap to the teacher to 3.9 points on average, while supervised recovery stays more than 20
points below. At matched compression, width pruning yields higher success and depth pruning
lower latency. On a 6-DoF manipulator, the
distilled student at 72\% reduction reaches 77.5\% success against 59.5\%
for supervised recovery, runs $2.23\times$ faster on-board than the teacher,
and uses 62\% less memory.
\end{abstract}

\section{Introduction}

A vision-language-action (VLA) model turns a camera image and a sentence into
robot motion. The leading open-source VLAs do this by attaching an action head
to a language backbone of several billion parameters
\cite{openvla,oft,cogact,pi0}. That backbone supplies the language and visual representations the policy acts
on. It also dominates the memory and the latency of a policy
query, so size and speed are now the main obstacles to on-board deployment.
Structured pruning reduces this cost directly and is well established for
language models \cite{llmpruner,minitron}. On a VLA, however, aggressive
pruning does not simply degrade the policy. It stops the policy from completing
the task. Removing 63\% of the effective language-backbone parameters of
OpenVLA-OFT lowers LIBERO-Long success from 93.2\% to 0.8\%, and recovery
determines whether the compressed policy is usable.

RLRC \cite{rlrc} recovers a 90\%-pruned model of this backbone to its dense
LIBERO score in two stages. Supervised fine-tuning recovers most of the lost success but
leaves a residual gap to the dense model, which a reinforcement-learning stage
then closes. That second stage needs a simulator or a robot in the loop, a
reward to optimize, and about 320 GPU-hours of training \cite{rlrc}. The dense
model itself remains available after pruning, and in language-model pruning
supervised recovery is routinely paired with distillation from it
\cite{minitron,hinton}. How much a pruned VLA can recover from its own teacher
without entering an environment is therefore an open question.

\begin{figure*}[t]
\centering
\includegraphics[width=\textwidth]{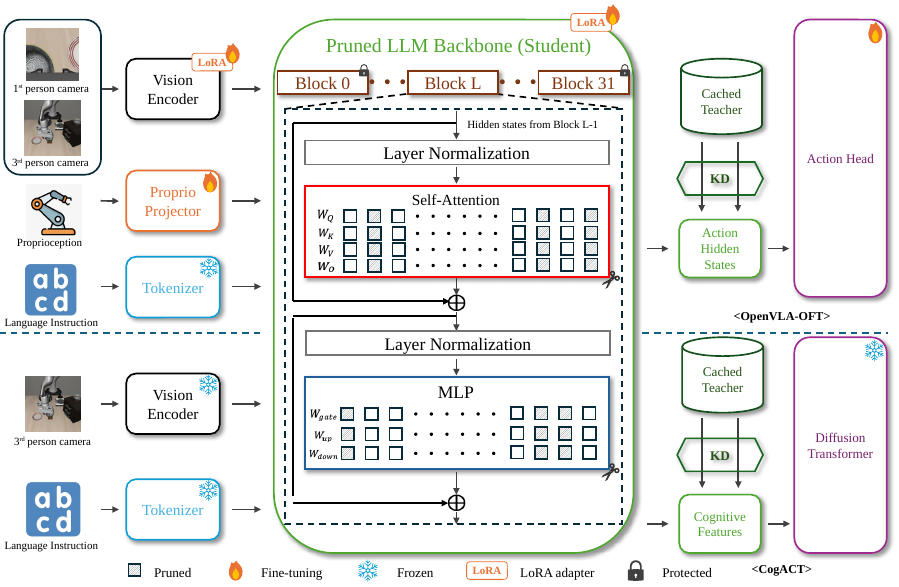}
\caption{Method overview. Width pruning removes whole attention heads and MLP
channels from every decoder block except the first and the last, which are
marked with lock icons. The teacher cache supplies the KD target, the action-token hidden states on
OpenVLA-OFT (top) and the cognition feature on CogACT (bottom).}
\label{fig:overview}
\end{figure*}

We eliminate most of this gap offline by applying a pruning method designed for this setting. Specifically, we prune attention heads and MLP channels while keeping the residual stream at its original width, so the teacher and student hidden states retain identical shapes. This allows the student to be trained to match the teacher’s states directly, with no projector needed, turning recovery into a supervised learning task. A single pass over the recovery dataset caches the inputs, the teacher’s hidden states, and the action targets, and the student trains solely from this cache (Fig.~\ref{fig:overview}). In practice, this restores most of the lost success in roughly 8 GPU-hours, and the benefit of hidden-state matching increases as the compression ratio rises (Section~\ref{sec:when}).

Determining when to stop compressing is a separate issue, and prior work typically reports only one or two ratios \cite{rlrc,gluestick,redundancy}. We sweep nine compression ratios, spanning 27\% to 89\% effective reduction, and consider two recovery objectives, using three seeds for each setting. The two backbones employ different action heads: the L1 regression head in OpenVLA-OFT and the diffusion transformer in CogACT. This sweep (Fig.~\ref{fig:curves}) indicates where the objectives diverge and where recovery begins to saturate. We then perform a controlled comparison of width vs. depth pruning and evaluate on a physical robot to convert these curves into a selected operating point.

\subsection*{Contributions}
\begin{itemize}\setlength{\itemsep}{1.5pt plus 1pt}
\item \textbf{Fully offline recovery of aggressively pruned VLAs.} Width
pruning leaves teacher and student hidden states directly comparable, which
turns recovery into offline supervision from a cached teacher, with no
rollouts and no reward.

\item \textbf{When hidden-state distillation is beneficial.} A sweep over nine
compression ratios and two backbones locates the reduction at which
hidden-state distillation begins to outperform supervised recovery, and shows
that a larger recovery budget moves that point further out.

\item \textbf{Width versus depth at matched compression.} With both choices
under one recovery protocol, width pruning yields higher success than a
CKA-guided depth baseline at every measured point on both backbones, and depth
pruning yields lower latency.

\item \textbf{Validation on a physical robot.} On a 6-DoF manipulator the
distilled student outperforms supervised recovery by 18.0 points and runs
$2.23\times$ faster on-board than its teacher with 62\% less memory.
\end{itemize}

\section{Related Work}

\textbf{Vision-language-action models.} Modern manipulation policies attach an
action decoder to a pretrained vision-language model (VLM)
\cite{openvla,pi0,rt2}. We build on the Prismatic VLM family \cite{prismatic}.
OpenVLA-OFT \cite{oft} pairs a Llama-2 7B backbone with a continuous L1 head
emitting 8-step action chunks. CogACT \cite{cogact} keeps the same VLM family
and replaces the head with a diffusion transformer \cite{dp,ddim}.

\textbf{Compressing VLAs.} Most efficiency work on VLAs \cite{survey} leaves the
weights unchanged. Token pruning discards uninformative visual tokens at
inference time \cite{tokenprune}, and action-space methods compress the action
representation so that the backbone need not be modified \cite{quart}. These methods act on the
inputs and the outputs, not on the weights themselves, and training compact
VLAs from scratch \cite{tinyvla,smolvla} is complementary to post-training
compression. For structural compression at aggressive ratios,
RLRC \cite{rlrc} is the closest prior work. It prunes the same backbone at a
90\% ratio and recovers it with supervised fine-tuning followed by
reinforcement learning. GLUESTICK \cite{gluestick} restores a pruned VLA
without training by interpolating dense and pruned weights, and a study \cite{redundancy} prunes 12--30\% without recovery. We target the ratios at
which the pruned policy no longer completes the task and training-based
recovery becomes necessary. Concurrent work removes transformer layers instead of channels, selected by
centered kernel alignment (CKA) similarity \cite{fewerlayers,cka} or by gate
sensitivity \cite{dropthen}. Both find that a substantial fraction of the
language blocks can be removed with supervised fine-tuning alone.
Section~\ref{sec:widthdepth} places the two pruning choices under a matched
recovery budget.

\textbf{Structured pruning and distillation in language models.} Taylor-based
importance scoring with LoRA recovery \cite{llmpruner,lora} is established for
language models, as are pipelines that combine pruning with distillation, such
as Minitron \cite{minitron}. The latter also finds width pruning better than
depth pruning at matched size. For large vision-language
models on VQA benchmarks, \cite{lvlmprune} draws the same conclusion on the
pruning choice and finds supervised fine-tuning plus hidden-state distillation
the best recovery. These findings come from single-step prediction. A policy is judged over a
horizon, where an error at one step changes every input that follows. We establish whether these findings still hold in closed-loop control, how much these techniques recover without environment interaction, and how their behavior changes with compression.

\section{Method}

At the target ratios, recovery needs to bring back most of the policy, and the dense teacher is the model that already performs the task successfully. We co-design pruning and recovery so this teacher can be used offline, with representations that are directly comparable to the student’s.

\subsection{Hidden-Preserving Structured Width Pruning}

The residual stream carries the representation that the action head reads.
Keeping that stream at its original width allows the teacher to be matched
state by state, without a projection layer. We therefore prune the language
backbone only, and leave the vision encoders, the action head, the embeddings,
the residual width, and the block count untouched.
Within each decoder block we remove whole attention heads and whole MLP
channels, so the pruned network stays a smaller dense model
(Fig.~\ref{fig:overview}). Unlike depth pruning, every block retains some
capacity, and the student starts recovery from a policy that still runs end to
end (Section~\ref{sec:widthdepth}). Following structured-pruning practice
\cite{llmpruner}, we exclude the first and last blocks, which sit next to the
embedding and output interfaces. The residual width and the block count are
the same at every ratio, so teacher and student hidden tensors have identical
shapes at corresponding layers. The distillation loss is then an element-wise
mean-squared error between them, with no projection to learn between the two
models.

The two excluded blocks make the nominal pruning ratio differ from the
effective parameter reduction. A nominal ratio of 50\% reduces the
backbone from 6.74\,B to 3.70\,B parameters, a reduction of 45.1\%. Compression levels refer
to effective reduction throughout.

\subsection{Action-Path Taylor Importance}
\label{sec:crit}

Importance measures the expected increase in loss if a given group is removed. We approximate this using the first-order Taylor term $|w\cdot g|$, aggregated across a set of calibration transitions and then added up within each head or channel group. The groups with the smallest scores are pruned in a single step, avoiding the expense of repeated re-estimation.

The loss that supplies the gradient determines which groups are preserved.
OpenVLA-OFT keeps a language-model head and an action-token vocabulary from
its base checkpoint, so the pretraining cross-entropy can still be computed.
That loss does not produce actions at inference, however, and a head the
policy relies on for action prediction can score low under it. We take the
gradient through the action head used at inference, the L1 regression loss on
OpenVLA-OFT and the noise-prediction loss on CogACT. Importance then approximates the increase in the action loss that
removing a group would cause. We call the two
criteria \emph{action Taylor} and \emph{CE-token Taylor} and compare them,
together with magnitude and random pruning, in Section~\ref{sec:norl}. This choice also constrains the recovery stage. Recovery adapts the groups
that remain but cannot restore a removed head, so the criterion fixes the
structure the student keeps for the rest of training. Two students that reach
the same success after recovery can still retain different groups, and that
difference becomes visible outside the recovery distribution.

\subsection{Offline Recovery from a Teacher Cache}
\label{sec:cache}

Running the teacher inside the training loop costs memory and recomputes the
same targets at every epoch, so we run the teacher once. A single offline pass
over
the recovery data stores the teacher's hidden states at the action-token
positions together with the ground-truth actions (Fig.~\ref{fig:overview}).
The student is then trained against this cache alone. Because inference is
deterministic and both models receive identical preprocessed inputs, the cache
holds the same targets a teacher inside the loop would produce. Caching also fixes
the target across epochs, so the teacher signal is identical across
objectives, ratios, and seeds. Any difference between SFT and
SFT+KD at a given ratio is then attributable to the objective alone.

Recovery adapts the student with LoRA on the language backbone and the vision
path, with the action head fully fine-tuned. The objective is
\begin{equation}
\mathcal{L} = \underbrace{\mathcal{L}_{\text{task}}(a_S, a_{\text{GT}})}_{\text{supervised}}
\; + \; \lambda \cdot \underbrace{\mathrm{MSE}(H_S, H_T)}_{\text{hidden-state KD}},
\end{equation}
where $\mathcal{L}_{\text{task}}$ is the L1 action loss on OpenVLA-OFT and the
diffusion noise-prediction loss on CogACT. We write SFT for supervised fine-tuning on the cached data alone and SFT+KD
when the hidden-state distillation term is added. To keep the KD term from
dominating the action loss at the start of recovery, $\lambda$ is ramped
linearly from zero over the early steps.

\textbf{Choice of distillation target.} The teacher can also be matched at the
action level, by mixing its predicted action into the supervised target. The
action output is far lower-dimensional than the representation that produces
it, so two students can agree on the action while their internal states
differ. The action loss supervises the representation only through that narrow
output. Hidden-state distillation directly supervises 56 action-token hidden states in OpenVLA-OFT and a single cognition vector in CogACT, covering many more dimensions per sample. The value of this extra supervision should grow with how much pruning disturbs
the representation, which the sweep over compression ratios measures. Both targets are compared in
Section~\ref{sec:when}.

The teacher is absent from the training loop, so recovery runs on a single
GPU. The cached observations are used without image augmentation, so student and
teacher inputs stay identical.

\section{Experimental Setup}
\label{sec:setup}

The benchmarks are LIBERO \cite{libero} and SimplerEnv (Google-Robot)
\cite{simplerenv}, on which the teacher models attain 93.2\% and 67.2\%
success. OpenVLA-OFT recovers on 101{,}468 cached LIBERO-Long transitions for
5{,}000 steps with $\lambda = 3$, about 8 GPU-hours on one H100, and is
evaluated over 500 episodes. CogACT recovers on 100{,}000 cached Google-Robot
transitions for 1{,}200 steps with $\lambda = 2$, about 1.7 GPU-hours on one
A6000, and is evaluated over 864 episodes.

Importance is scored over 512 calibration transitions. The student is adapted
with LoRA of rank 32 and $\alpha = 16$, applied to the language backbone and
the vision path on OpenVLA-OFT and to the language model only on CogACT. The
action head is fully fine-tuned on OpenVLA-OFT and frozen on CogACT. The ramp
on $\lambda$ covers the early steps of recovery on both backbones. Every
recovery run requires a single 48\,GiB GPU.

\textbf{Benchmarks.} The sweep uses LIBERO-Long, and the recovery procedure is
evaluated on all four LIBERO suites (Section~\ref{sec:norl}). CogACT is
evaluated on all four SimplerEnv task families. Robustness is measured on LIBERO-Plus \cite{liberoplus}, which
perturbs LIBERO tasks along seven dimensions, over 2{,}519 episodes paired
with the teacher.

\textbf{Paired evaluation.} Student and teacher are evaluated from an identical
set of initial states, so that each episode is matched one-to-one. On
OpenVLA-OFT the paired episodes are scored with McNemar's exact test on the
discordant pairs. The 864 CogACT episodes are 216 initial states with four
texture variants each, so the paired test is clustered by initial state. The test is a one-sample $t$-test on the 216 cluster means. Every contrast is
evaluated separately for each training seed at the 0.05 level.

\textbf{Seeds.} All results are given for the final checkpoint, without
checkpoint selection. Every recovered point of the sweep is evaluated with
three seeds, except two OpenVLA-OFT points evaluated with six (63\%
reduction) and five (85\%), and seed means are given throughout. Models pruned without recovery are evaluated once per point, and the control experiments use
three seeds unless their figure states otherwise.

\textbf{CogACT evaluation subset.} The released checkpoint does not reproduce
the published result on the put-in-drawer family \cite{cogactissue}, so paired
comparisons against the teacher use the remaining 756 episodes. Success rates
are reported over all 864, and the figures plot the mean across seeds.

\begin{table*}[t]
\centering
\caption{Success rate (\%) after width pruning at nine compression ratios,
with and without recovery. $\Delta$ denotes SFT+KD minus SFT in percentage
points (bold, $^{*}p<0.05$, $^{**}p<0.01$, $^{***}p<0.001$). Three seeds per
point, six at 63\% and five at 85\% on OpenVLA-OFT. On OpenVLA-OFT success
without recovery already reaches zero at 72\%, and higher ratios are marked
n/a.}
\label{tab:grid}
\renewcommand{\arraystretch}{0.95}
\setlength{\tabcolsep}{5pt}
\begin{tabular}{cc cccc cccc}
\toprule
\multicolumn{2}{c}{Reduction (\%)} &
\multicolumn{4}{c}{OpenVLA-OFT / LIBERO-Long (500 ep)} &
\multicolumn{4}{c}{CogACT / SimplerEnv (864 ep)} \\
\cmidrule(lr){1-2}\cmidrule(lr){3-6}\cmidrule(lr){7-10}
Nominal & Effective & none & SFT & SFT+KD & $\Delta$ &
                      none & SFT & SFT+KD & $\Delta$ \\
\midrule
30 & 27 & 89.2 & 90.2 & 91.9 & $+1.7$ & 67.8 & 75.8 & 75.2 & $-0.6$ \\
50 & 45 & 54.4 & 89.9 & 90.0 & $+0.1$ & 61.5 & 72.0 & 74.9 & $+2.9$ \\
70 & 63 & 0.8  & 87.6 & 89.7 & $\mathbf{+2.1^{**}}$  & 43.9 & 61.6 & 71.9 & $\mathbf{+10.3^{***}}$ \\
80 & 72 & 0.0  & 87.9 & 90.1 & $\mathbf{+2.2^{*}}$   & 21.4 & 46.5 & 68.6 & $\mathbf{+22.1^{***}}$ \\
90 & 81 & n/a  & 84.1 & 88.6 & $\mathbf{+4.5^{***}}$ & 5.7  & 36.6 & 56.2 & $\mathbf{+19.6^{***}}$ \\
95 & 85 & n/a  & 82.6 & 85.0 & $\mathbf{+2.4^{**}}$  & 2.3  & 29.0 & 45.9 & $\mathbf{+16.9^{***}}$ \\
97 & 87 & n/a  & 78.6 & 82.1 & $\mathbf{+3.5^{**}}$  & 1.9  & 34.5 & 43.9 & $\mathbf{+9.4^{***}}$ \\
98 & 88 & n/a  & 74.8 & 77.1 & $+2.3$ & 0.9 & 29.7 & 44.8 & $\mathbf{+15.1^{***}}$ \\
99 & 89 & n/a  & 76.7 & 74.7 & $-2.0$ & 1.6 & 31.9 & 45.8 & $\mathbf{+13.9^{***}}$ \\
\midrule
\multicolumn{2}{c}{teacher} & \multicolumn{4}{c}{93.2} & \multicolumn{4}{c}{67.2} \\
\bottomrule
\end{tabular}
\end{table*}

\begin{figure*}[t]
\centering
\includegraphics[width=\textwidth]{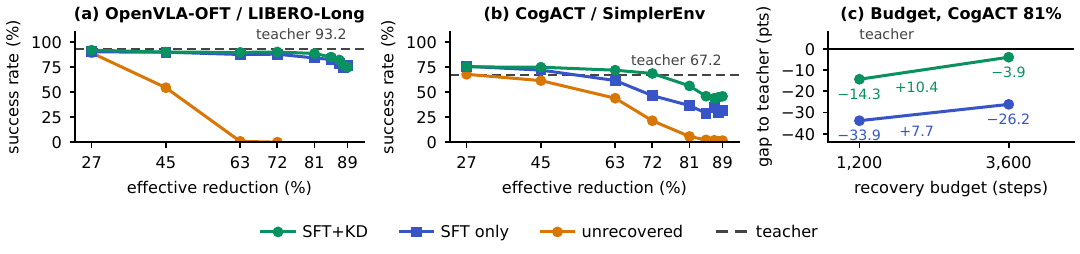}
\caption{Success against effective reduction. (a) OpenVLA-OFT on LIBERO-Long
and (b) CogACT on SimplerEnv, on identical axes; recovery curves are seed
means. (c) Gap to the teacher at 81\% reduction on CogACT for the 1{,}200-step
and 3{,}600-step budgets; zero denotes teacher success.}
\label{fig:curves}
\end{figure*}

\section{Results}

\subsection{Offline Recovery Without Reinforcement Learning}
\label{sec:norl}

Without recovery, performance falls quickly with compression, from 89.2\% at
27\% effective reduction to 54.4\% at 45\%, 0.8\% at 63\%, and no successful
episode at 72\%. Only the Taylor-based criteria retain nonzero success at moderate ratios
(Fig.~\ref{fig:controls}(a)). At 45\% reduction action Taylor retains 54.4\%,
against 50.2\% for CE-token Taylor and 0\% for magnitude and random pruning,
a significant margin of 4.2 points.
Magnitude is not conditioned on the action objective, and the cross-entropy
gradient reaches it only indirectly, so the ranking follows how
directly each criterion sees that objective.

Offline recovery reverses this collapse. Hidden-state distillation returns
every one of these checkpoints above 89\%, including those that complete no
episode before recovery (Table~\ref{tab:grid}). The 63\%-reduced model, at 0.8\% before
recovery, attains $89.7\pm2.4\%$ against a teacher at 93.2\%, a recovery of
88.9 points. Supervised recovery alone attains 87.6\%, and the KD term reduces the remaining 5.6-point gap by 2.1 points.

\textbf{Comparison with RL-based recovery.} RLRC \cite{rlrc} finds
the same pattern on this backbone at a 90\% pruning ratio. LIBERO-Long falls
from 94.5\% to 0\% and supervised recovery returns it to 86.8\%. A
reinforcement-learning stage removes the remainder, reaching 94.8\%. At the
same nominal ratio our supervised baseline shows a residual gap of similar
size, reaching 84.1\% against a 93.2\% teacher. The KD term recovers half of
what remains, reaching 88.6\% from the dense teacher alone. Our recovery
takes about 8 H100 GPU-hours and no rollouts, while RLRC reports about 320
GPU-hours for its recovery pipeline with a simulator in the loop.

Recovery narrows both the gap to the teacher and the differences among
criteria. On OpenVLA-OFT, the 27\% SFT+KD student performs on par with its teacher, and
from 45\% reduction onward every recovered student stays within a few points
of it (Table~\ref{tab:grid}). Magnitude- and random-pruned
students at 45\% reduction recover from 0\% to 90.5\% and 87.4\% under the
same procedure. The CE-token student recovers to $90.0\pm1.7\%$, the same
success as the $90.0\pm1.6\%$ of action Taylor, so the two criteria are
indistinguishable after recovery. CogACT shows the same pattern. At 85\% reduction, where pruning leaves the
model at 2.3\%, recovery lifts it to 45.9\%. At 45\% reduction the student scores 74.9\% on
all 864 episodes against a 67.2\% teacher. On the 756-episode subset it exceeds the teacher by 3.0 points. The same procedure generalizes to the remaining LIBERO suites without any hyperparameter changes. At 45\% reduction the spatial, object, and goal suites reach
97.5\%, 96.0\%, and 97.0\%, against teachers at 98.2\%, 97.2\%, and 97.4\%.

The differences among criteria grow again under distribution shift. On
LIBERO-Plus, after identical recovery, the action Taylor student reaches
63.4\%, against 58.6\% for magnitude and 50.1\% for random pruning. Recovery can refit the groups that remain on the
recovery distribution, so what distinguishes the criteria here is the set of
groups each one retained. Scoring through the action head keeps the groups the
policy uses to act, and that choice is not recoverable later
(Section~\ref{sec:crit}).

\subsection{When Hidden-State Distillation Is Beneficial}
\label{sec:when}

The recovery objective becomes relevant only beyond moderate compression, so
the comparison between the two objectives depends on the ratio at which it is
made. Up
to 45\% reduction the objective makes no significant difference. On OpenVLA-OFT the two
recovery objectives differ by $+1.7$ points at 27\% and $+0.1$ points at 45\%,
neither significant on any seed, and action-level matching and its combination with hidden-state
distillation fall in the same 88.4--90.0\% band. On CogACT the differences are
$-0.6$ and $+2.9$ points, neither significant (Table~\ref{tab:grid}).

The two objectives first separate significantly at 63\% reduction, in the same direction on
both backbones, by a significant $+2.1$ points on OpenVLA-OFT and $+10.3$
points on CogACT. On CogACT the advantage of distillation is largest between 72\% and 81\%
reduction, where it reaches $+19.6$ to $+22.1$ points and is significant on
every seed. On OpenVLA-OFT it stays between $+2.1$ and $+4.5$ points from 63\%
to 87\%.

The advantage holds over a wide range of compression, and its extent differs
between the two backbones. On CogACT it remains significant through 89\%
reduction, and on OpenVLA-OFT through 87\%. 
Beyond roughly 85\% reduction the backbones themselves diverge, with CogACT
plateauing near 44--46\% and OpenVLA-OFT declining from 85.0\% to 74.7\%.

\begin{figure*}[t]
\centering
\includegraphics[width=\textwidth]{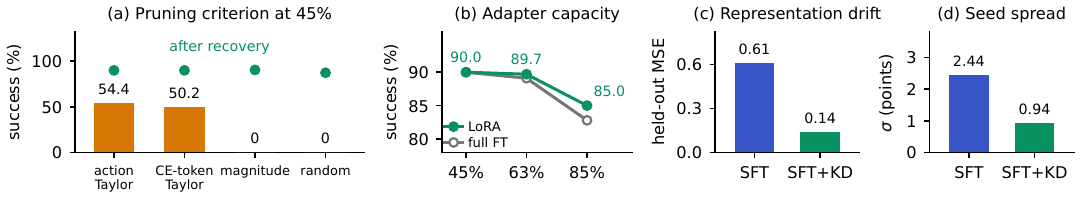}
\caption{Control experiments. (a) Success before recovery by pruning
criterion at 45\% reduction; the dots show the same models after hidden-state
KD recovery. (b) LoRA (green, filled markers) versus full fine-tuning (gray,
open markers), with three, two, and one seeds at 45\%, 63\%, and 85\%. (c) Held-out cognition-feature MSE to the teacher under each
objective. (d) Run-to-run spread at 72\% reduction across seeds.}
\label{fig:controls}
\end{figure*}

The two objectives preserve different properties. Supervised recovery fits the
task but allows the student's representation to drift from the teacher's,
whereas the KD term constrains it. On episodes never
seen during recovery, the SFT student's mean-squared error to the teacher's
cognition feature on CogACT is 0.61, against 0.14 for the SFT+KD student
(Fig.~\ref{fig:controls}(c)). The action loss reaches the representation only through the action output, so
the more pruning disturbs that representation, the less the supervised term
constrains it. This
is why the two
objectives separate only beyond moderate compression, and why more steps alone
do not close the difference (Section~\ref{sec:budget}). The KD term also reduces run-to-run variability at high compression. At 72\% reduction the seed spread is 2.44 points
for SFT against 0.94 for SFT+KD (Fig.~\ref{fig:controls}(d)).

Full fine-tuning with the same objective attains 90.0\%, 89.1\%, and 82.8\%
at 45\%, 63\%, and 85\% reduction. These lie within the seed range of the LoRA
runs (Fig.~\ref{fig:controls}(b)), so LoRA performs on par with full
fine-tuning at these points and the recovery cost remains at the adapter
level.

\subsection{Recovery Limits}
\label{sec:budget}

The recovered student remains close to its teacher over a wide range of
compression. On the 756-episode subset, at the standard 1{,}200-step budget, the student
tracks the teacher up to 72\% reduction, falling at most 4.4 points below it
and never by a significant margin. At 81\% the gap grows to 12 to 18
points, and depends on the training budget. Tripling that budget to 3{,}600
steps narrows the gap to the teacher from 14.3 to 3.9 points on average
($-1.1$, $-2.0$, and $-8.6$ points across three seeds; not significant on two
of three, $p = 0.66$, $0.34$, and $7.9\times10^{-4}$), and improves success by
10.4 points on average, significant on two of three seeds. The reachable compression is set by the recovery procedure and its budget, not
by the pruned architecture alone.

The budget alone does not close the gap at 81\%. If the two objectives
differed only in how fast they converge, a matched budget would close it. Extending supervised-only recovery at 81\% reduction to the same
3{,}600-step budget lifts it significantly, by about 8 points, yet that
student finishes more than 23 points below the teacher. Hidden-state KD at that budget finishes
3.9 points below on average (Fig.~\ref{fig:curves}(c)). The separation between the objectives is as large at the matched budget as at
1{,}200 steps. The gain from the larger budget depends on the objective, and the two act together instead of substituting for each other.

Past 85\% reduction, success plateaus around 44--46\% at the 1{,}200-step
budget. At 81\% reduction on CogACT, varying $\lambda$ five-fold moves success by at
most 0.5 points, within 57.1--57.6\%. Tripling the budget changes success by more than
ten points.

\begin{figure*}[t]
\centering
\includegraphics[width=\textwidth]{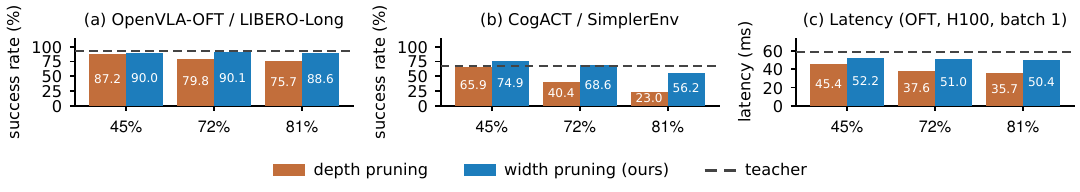}
\caption{Depth versus width pruning at matched effective reduction, both
recovered with hidden-state KD. (a) OpenVLA-OFT on LIBERO-Long and (b) CogACT
on SimplerEnv, three seeds per point; the dashed line marks the teacher.
(c) Latency of the OpenVLA-OFT students (H100, batch 1).}
\label{fig:widthdepth}
\end{figure*}

\subsection{Width versus Depth Pruning}
\label{sec:widthdepth}

Depth and width pruning have so far been developed with their own recovery
pipelines, so we place both under one protocol and vary only the pruning
choice. We reimplement the
CKA-guided layer selection of \cite{fewerlayers} and evaluate it under the
same recovery protocol, budget, evaluation procedure, and statistics.
Depth- and width-pruned models are matched by effective reduction at three
points on both backbones.

Under this controlled setting, width pruning achieves higher success than the
CKA-guided depth baseline at every measured point on both backbones, with
every contrast significant (Fig.~\ref{fig:widthdepth}). The margin is 2.8 to
12.9 points on OpenVLA-OFT and 9.0 to 33.2 points on CogACT.
The margin is present from the mildest ratio we test and grows with
compression.

Depth and width pruning differ qualitatively before recovery. Depth-pruned models
reach 0\% success at all three OpenVLA-OFT points, while width pruning at
45\% retains 54.4\%, so recovery begins from a functioning policy. A removed layer eliminates every computation it performed, while a
narrowed layer retains the highest-scoring heads and channels at every depth. A similar ordering has been found for language models \cite{minitron} and
vision-language models \cite{lvlmprune}, and the matched protocol here shows
that the same ordering holds under closed-loop control.

On latency, the ordering reverses. Depth pruning is faster at every measured
point, by $1.31\times$ to $1.66\times$ against $1.14\times$ to $1.18\times$
for width. Dropping a layer removes sequential computation, while narrowing one leaves
the number of sequential steps unchanged.
Depth-pruned models also train about 22\% faster per step, consistent with the
training-cost advantage found in \cite{fewerlayers}. Width pruning and depth pruning thus present a trade-off, one preserving task
success and the other yielding larger latency reductions.

On CogACT, where the target is a single cognition vector, the hidden-state KD
improves success at every depth ratio by 7.8 to 22.0 points. On OpenVLA-OFT,
where the target is spread over 56 action-token positions, the gain is
significant at one of three points. A target spread over many positions is harder to reproduce once whole
layers are missing, so the distillation target and the pruning choice do not
act independently.

\subsection{Deployment Measurements}
\label{sec:deploy}

Latency does not follow the parameter count (Table~\ref{tab:deploy}). On
CogACT latency falls from 129.1\,ms to 85.0\,ms at 72\% reduction and then
stays at
84.7--85.0\,ms.
The diffusion sampler runs ten DDIM steps independently of backbone size, and
that fixed cost sets a latency floor that further narrowing cannot lower.
OpenVLA-OFT latency stays within 50--52\,ms at every ratio, consistent with
the unchanged layer count limiting further reductions
(Section~\ref{sec:widthdepth}). Memory follows the retained parameter count on
both backbones, so latency and memory must be read separately.

\begin{table}[t]
\centering
\caption{Deployment at batch 1. CogACT runs on an A6000 in BF16 with an FP32
diffusion head and ten DDIM steps at guidance scale 1.5, and OpenVLA-OFT on an
H100 in BF16. Backbone denotes the retained backbone size and success rates are those of
Table~\ref{tab:grid}. Latency is the minimum over repeated batches of 100
forward passes after warm-up, and speed is the teacher-to-student latency
ratio.}
\label{tab:deploy}
\renewcommand{\arraystretch}{0.95}
\setlength{\tabcolsep}{4pt}
\begin{tabular}{lccccc}
\toprule
Model & Backbone & SR (\%) & Latency (ms) & Speed & VRAM (GiB) \\
\midrule
\multicolumn{6}{c}{\emph{CogACT / SimplerEnv}} \\
teacher      & 6.74\,B & 67.2 & 129.1 & $1.00\times$ & 14.7 \\
45\% KD  & 3.70\,B & 74.9 & 97.3  & $1.33\times$ & 9.1 \\
63\% KD  & 2.51\,B & 71.9 & 89.4  & $1.44\times$ & 6.8 \\
72\% KD  & 1.86\,B & 68.6 & 85.0  & $1.52\times$ & 5.6 \\
81\% KD  & 1.26\,B & 56.2 & 84.7  & $1.52\times$ & 4.4 \\
\midrule
\multicolumn{6}{c}{\emph{OpenVLA-OFT / LIBERO-Long}} \\
teacher      & 6.74\,B & 93.2 & 59.3 & $1.00\times$ & 15.9 \\
45\% KD  & 3.70\,B & 90.0 & 52.2 & $1.14\times$ & 10.2 \\
63\% KD  & 2.51\,B & 89.7 & 51.1 & $1.16\times$ & 7.9 \\
72\% KD  & 1.86\,B & 90.1 & 51.0 & $1.16\times$ & 6.8 \\
81\% KD  & 1.26\,B & 88.6 & 50.4 & $1.18\times$ & 5.6 \\
\bottomrule
\end{tabular}
\end{table}

These measurements identify 72\% reduction as the operating point. At that
ratio, and at the 1{,}200-step budget, the CogACT student scores 68.6\% on all
864 episodes
against a 67.2\% teacher, matching it on the 756-episode subset. It runs $1.52\times$ faster and uses 62\% less memory at no cost in success.
Further compression saves memory but not latency. Locating such a point requires a sweep of this density, since one or two
ratios do not show where success, latency, and memory cease to move together.

\textbf{Robustness under distribution shift.} The advantage of the KD term
persists under perturbation. Over the 1{,}992 SimplerEnv variant
episodes on CogACT,
the 72\% SFT+KD student reaches 60.7\% against 37.2\% for the SFT student and
66.6\% for the teacher. The margin of 23.5 points is as large as the $+22.1$
points measured in distribution (Table~\ref{tab:grid}). On
LIBERO-Plus the recovered 45\% OpenVLA-OFT student stays within 1.8 points of
its teacher, at 63.4\% against 65.2\% over 2{,}519 initial states.

\subsection{Real-Robot Validation}
\label{sec:robot}

\begin{figure}[t]
\centering
\includegraphics[width=\columnwidth]{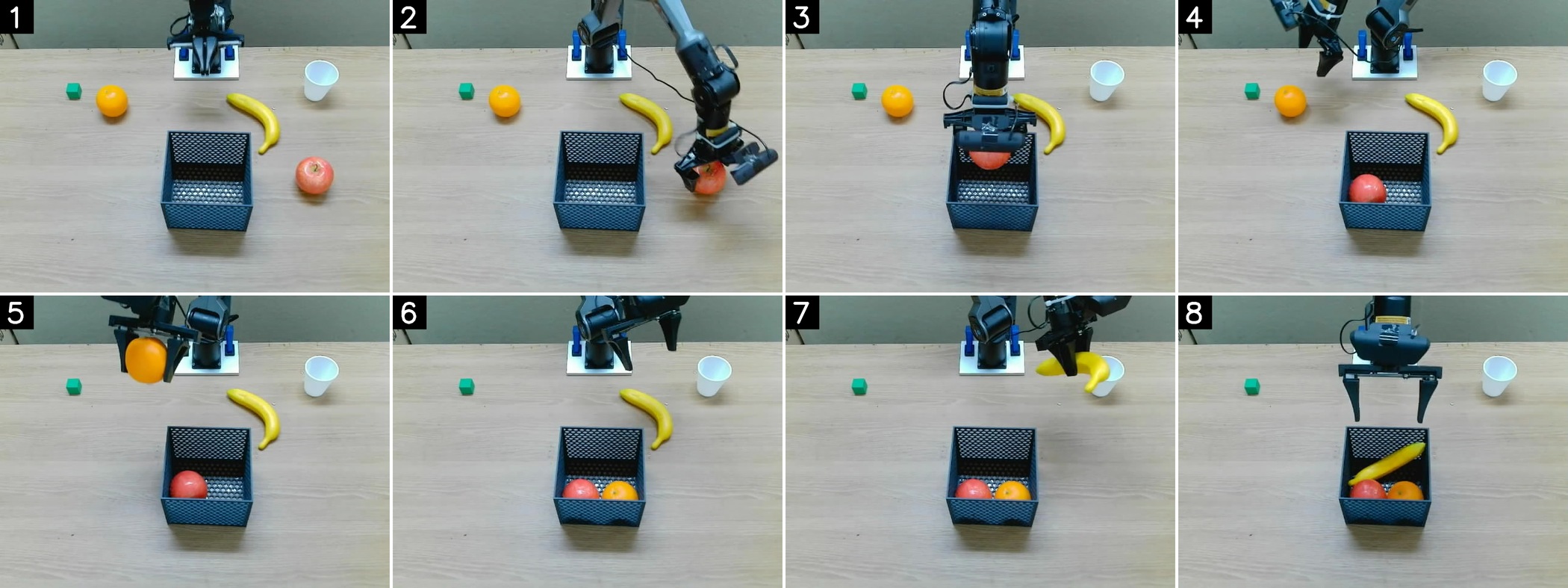}
\caption{A successful rollout of the 72\% SFT+KD student on a long-horizon
task. Frames are ordered from left to right and top to bottom. Three objects are
placed into the basket in sequence (1--4 tomato, 5--6 orange, 7--8 banana);
the green cube and the cup are distractors.}
\label{fig:robot}
\end{figure}

On an AgileX PiPER 6-DoF arm (Fig.~\ref{fig:robot}) we evaluate the teacher
and the 72\% OpenVLA-OFT students recovered with SFT and with SFT+KD. The two
students share the architecture, the cache, and the budget, and differ only in
the recovery objective. The teacher is fine-tuned on 450 demonstrations collected across 10 tasks. The
students are recovered from a cache built over the same data and trained for
20{,}000 steps, with the adapter configuration of Section~\ref{sec:setup}. Each model is run for 20 trials per
task over the same 20 layouts and scored with McNemar's exact test on the
paired episodes. Latency and memory are measured on the robot's on-board
Jetson Thor (Table~\ref{tab:robot}).

\begin{table}[t]
\centering
\caption{Real-robot validation on an AgileX PiPER 6-DoF arm. The columns are
the OpenVLA-OFT teacher and the two students pruned to 72\% effective
reduction; the last row pools the 200 episodes instead of averaging the three
families. Latency is the median of 50 repetitions of one full policy query on the
on-board Jetson Thor, measured after warm-up.}
\label{tab:robot}
\renewcommand{\arraystretch}{0.95}
\setlength{\tabcolsep}{4pt}
\begin{tabular}{lccc}
\toprule
 & & \multicolumn{2}{c}{72\% student} \\
\cmidrule(lr){3-4}
Task family & teacher & SFT & SFT+KD \\
\midrule
\multicolumn{4}{c}{\emph{success rate (\%)}} \\
object (4 tasks, 80 ep)       & 78.8 & 78.8 & \textbf{90.0} \\
spatial (3 tasks, 60 ep)      & 56.7 & 50.0 & \textbf{70.0} \\
long-horizon (3 tasks, 60 ep) & 56.7 & 43.3 & \textbf{68.3} \\
overall (200 ep)              & 65.5 & 59.5 & \textbf{77.5} \\
\midrule
\multicolumn{4}{c}{\emph{on-board deployment}} \\
latency (ms)                  & 362 & 163 & \textbf{162} \\
memory (GiB)                  & 15.0 & 5.7 & \textbf{5.7} \\
\bottomrule
\end{tabular}
\end{table}

\textbf{Success.} The contrast between the two objectives also appears on
hardware. Over 200 paired episodes the SFT+KD student reaches 77.5\%, outperforming the
SFT student by 18.0 points ($p<0.001$). It also
exceeds the teacher's
65.5\% by 12.0 points ($p = 0.004$), whereas the same 72\% operating point in
simulation places the student 3.1 points below its teacher. The SFT student, by
contrast, remains 6.0 points below the teacher. Both students also receive the
ground-truth action targets during recovery, which the teacher did not. The
margin over SFT is largest on the long-horizon family ($+25.0$) and the spatial
family ($+20.0$), both significant, and smallest on the object family
($+11.2$, not significant). Longer tasks give errors more steps to compound,
so remaining close to the teacher's representation matters most there.

\textbf{Memory and latency.} On-board memory drops from 15.0\,GiB to
5.7\,GiB, a 62\% saving, close to the drop from 15.9 to 6.8\,GiB measured on
the workstation (Table~\ref{tab:deploy}). Both students are structurally identical, so they share that footprint and
differ in latency by 1\,ms. Hidden-state distillation adds no parameters or
computation at inference.
On-arm latency falls from 362\,ms to 162\,ms, a speedup of $2.23\times$ and
about twice the $1.16\times$ measured on the H100 at the same ratio. The same
compression therefore yields more speed on the robot's own hardware.

\section{Discussion}

The sweep provides a direct account of how far a VLA can be compressed and of
what recovery it requires, on a single GPU, with no simulator and no reward. Up to 45\% reduction, supervised fine-tuning on the cached
recovery data recovers about as much as distillation does, with no significant
difference between the two, and the KD term can be
omitted. Beyond that point, supervised recovery alone leaves a larger
gap to the teacher's representation. Matching hidden states limits that drift, which improves
success by more than 20 points on CogACT at high compression.

The compression ratio should be chosen together with the recovery budget,
since the two jointly determine the attainable success. A larger budget moves
the student closer to the teacher, and under the distillation objective it
reduces a gap of more than 12 points to about four on average.

Between the two pruning choices, width favors task success and depth favors
latency. On the Jetson Thor, width pruning yields about twice the speedup
measured on the workstation while keeping its success advantage. Most of the latency that width pruning sacrifices on a workstation is regained
on robot hardware.

\section{Limitations}

All compression ratios are swept in simulation, and the real-robot evaluation
covers the 72\% operating point over 200 paired episodes. Both backbones share
the Prismatic VLM family, so the results show consistent behavior across two
action-head paradigms, and extending the analysis to other VLM families is a
natural next step. Recovery uses fixed, unaugmented cached observations, and
alternative distillation targets were compared at one operating point.

\section{Conclusion}

A pruned vision-language-action model can be recovered from a cache of
demonstrations and teacher hidden states alone, without reinforcement
learning, rollouts, or a reward. Supervised recovery restores most of the lost
success, and hidden-state distillation adds a significant further gain between
63\% and 87\% reduction on OpenVLA-OFT and from 63\% to 89\% on CogACT. A
larger recovery budget narrows the gap of the student at 81\% reduction to
3.9 points on average.
The advantage also appears on a physical robot, where the
distilled student runs $2.23\times$ faster than its teacher and adds no memory
or latency over supervised recovery.

\end{document}